\documentclass{article}

\usepackage{times}
\usepackage{graphicx} 
\usepackage{subfigure} 

\usepackage{natbib}

\usepackage{algorithm}
\usepackage{algorithmic}
\usepackage{amsmath}

\usepackage{hyperref}

\usepackage[accepted]{icml2015}

\icmltitlerunning{Feedforward Sequential Memory Networks (FSMN)}

\begin{document} 

\twocolumn[
\icmltitle{Feedforward Sequential Memory Networks: \\ A New Structure to Learn Long-term Dependency }
\vskip 0.1in
\icmlauthor{Shiliang Zhang$^1$, Cong Liu$^2$, Hui Jiang$^3$, Si Wei$^2$, Lirong Dai$^1$, Yu Hu$^2$}{}
\icmladdress{
$^1$ NELSLIP, University of Science and Technology of China, Hefei, Anhui, China\\
$^2$ IFLYTEK Research, Hefei, Anhui, China \\
$^3$ Department of Electrical Engineering and Computer Science, York University, Toronto, Ontario, Canada \\
{\em Emails: zsl2008@mail.usct.edu.cn, \{congliu2,siwei,yuhu\}@iflytek.com, hj@cse.yorku.ca, lrdai@ustc.edu.cn}}
\vskip 0.3in
]

\begin{abstract} 
In this paper, we propose a novel neural network structure, namely \emph{feedforward sequential memory networks (FSMN)}, to model long-term dependency in time series without using recurrent feedback. The proposed FSMN is a standard  fully-connected feedforward neural network equipped with some learnable memory blocks in its hidden layers. The memory blocks use a tapped-delay line structure to encode the long context information into a fixed-size representation as short-term memory mechanism. We have evaluated the proposed FSMNs in several standard benchmark tasks, including speech recognition and language modelling. Experimental results have shown FSMNs significantly outperform the conventional recurrent neural networks (RNN), including LSTMs, in modeling sequential signals like speech or language. Moreover, FSMNs can be learned much more reliably and faster than RNNs or LSTMs due to the inherent non-recurrent model structure. 
\end{abstract} 

\section{Introduction} 

For a long time, artificial neural networks (ANN) have been widely regarded as an effective learning machine for self-learning feature representations from data to perform pattern classification and regression tasks. In recent years, as more powerful computing resources (e.g., GPUs) become readily available and more and more  real-world data are being generated, {\em deep learning} \cite{lecun2015deep,Juergen2015} is reviving as an active research area in machine learning during the past few years. The surge of deep learning aims to learn neural networks with a deep architecture consisting of many hidden layers between input and output layers, and thousands of nodes in each layer. The deep network architecture can build hierarchical representations with highly non-linear transformations to extract complex structures, which is similar to the human information processing mechanism (e.g., vision and speech). Depending on how the networks are connected, there exist various types of deep neural networks, such as feedforward neural networks (FNN) and  recurrent neural networks (RNN). 
 
FNNs are organized as a layered structure, including an input layer, multiple hidden layers and an output layer. The outputs of a hidden layer are a weighted sum of its inputs coming from the previous layer, then followed by a non-linear transformation. Traditionally, the sigmoidal nonlinearity, i.e., $f(x)=1/(1+e^{x})$, has been widely used. Recently, the most popular non-linear function is the so-called rectified linear unit (ReLU), i.e., $f(x)=\max(0,x)$ \cite{jarrett2009best,nair2010rectified}. 
In many real-word applications, 
it is experimentally shown that ReLUs can learn deep networks more efficiently, allowing to train a deep supervised network without any unsupervised pre-training. The two popular FNN architectures are fully-connected deep neural networks (DNN) and convolutional neural networks (CNN).  The structure of DNNs is a conventional multi-layer perceptron with many hidden layers, where units from two adjacent layers are fully connected, but no connection exists among the units in the same layer. On the other hand, inspired by the classic notions of simple cells and complex cells in visual neuroscience \cite{hubel1962receptive}, CNNs are designed to hierarchically process the data represented in the form of multiple location-sensitive arrays, 
such as images.
The uses of local connection, weight sharing and pooling make CNNs insensitive to small shifts and distortions in raw data. Therefore, CNNs are widely used in a variety of real applications, including document recognition \cite{lecun1998gradient}, image classification \cite{ciresan2011flexible, krizhevsky2012imagenet}, face recognition \cite{lawrence1997face,taigman2014deepface}, speech recognition \cite{abdel2012applying,sainath2013deep}.

When neural networks are applied to sequential data such as language, speech and video, it is crucial to model the long term dependency in time series. Recurrent neural networks (RNN) \cite{elman1990finding} are designed to capture long-term dependency within the sequential data using a simple mechanism of recurrent feedback. Moreover, the bidirectional RNNs \cite{schuster1997bidirectional} have also been proposed to incorporate the context information from both directions (the past and future) in a sequence.
RNNs can learn to model sequential data over an extended period of time and store the memory in the network weights, then carry out rather complicated transformations on the sequential data. RNNs are theoretically proved to be a turing-complete machine \cite{Siegelmann1995}. 
As opposed to FNNs that can only learn to map a fixed-size input to a fixed-size output, RNNs can in principle learn to map from one variable-length sequence to another. While RNNs are theoretically powerful, the learning of RNNs relies on the so-called back-propagation through time (BPTT) \cite{werbos1990backpropagation} due to the internal recurrent cycles. The BPTT significantly increases the computational complexity of the learning, and even worse, it may cause many problems in learning, such as gradient vanishing and exploding \cite{bengio1994learning}. Therefore, some new architectures have been proposed to alleviate these problems. For example, the long short term memory (LSTM) model \cite{hochreiter1997long,gers2000learning} is an enhanced RNN architecture to implement the recurrent feedbacks using various learnable gates, which ensure that the gradients can flow back to the past more effectively. LSTMs have yielded promising results in many applications, such as sequence modeling \cite{graves2013generating}, machine translation \cite{cho2014learning}, speech recognition \cite{graves2013speech,sak2014long} and many others. More recently, a simplified model called gated recurrent unit (GRU)  \cite{cho2014learning} is proposed and reported to achieve similar performance as LSTMs \cite{Chung2014Empirical}. Finally, in the past year, there are some latest research effort to use various forms of explicit memory units to construct neural computing models that can have longer-term memory \cite{Graves2014NTM,Weston2014MN}. For example,  the so-called neural turing machines (NTM) \cite{Graves2014NTM} are proposed to improve the memory of neural networks by coupling with external memory resources, which can learn to sort a small set of numbers as well as other symbolic manipulation tasks. Similarly, the memory networks \cite{Weston2014MN} employ a memory component that supports some learnable read and write operations.

Compared with FNNs, an RNN is deep in time so that it is able to capture the long-term dependency in sequences. Unfortunately, the high computational complexity of learning makes it difficult to scale RNN or LSTM based models to larger tasks. Because the learning of FNN is much easier and faster, it is somehow preferable to use a feedforward structure to learn the long-term dependency in sequences. 
A straightforward attempt is the so-called unfolded RNN \cite{saon2014unfolded}, where an RNN is unfolded in time for a fixed number of time steps. The unfolded RNN only needs comparable training time as the standard FNNs while achieving better performance than FNNs. However, the context information learned by the unfolded RNNs is still very limited due to the limited number of unfolding steps in time. 
Moreover, it seems quite difficult to derive an unfolded version
for more complex recurrent architectures, such as LSTM.

In this work, we propose a simple structure, namely \emph{feedforward sequential memory networks (FSMN)}, which can effectively model long-term dependency in sequential data without using any recurrent feedback.  The proposed FSMN is inspired by the filter design knowledge in digital signal processing \cite{oppenheim1989discrete} that any infinite impulse response (IIR) filter can be well approximated using a high-order finite impulse response (FIR) filter. Because the recurrent layer in RNNs can be conceptually viewed as a first-order IIR filter, it should be precisely approximated by a high-order FIR filter. Therefore, we extend the standard feedforward fully connected neural networks by augmenting some memory blocks, which adopt a tapped-delay line structure as in FIR filters, into the hidden layers. As a result, the overall FSMN remains as a pure feedforward structure so that it can be learned in a much more efficient and stable way than RNNs.  The learnable FIR-like memory blocks in FSMNs may be used to encode long context information into a fixed-size representation, which helps the model to capture long-term dependency. We have evaluated FSMNs on several benchmark tasks in the areas of speech recognition and language modeling, where RNNs or LSTMs currently excel at.  For language modeling tasks, the proposed FSMN based language models can significantly overtake not only the standard FNNs but also the popular RNNs and LSTMs by a significant margin. As for the speech recognition, experiments on the standard Switchboard (SWB) task show that FSMNs can even outperform the state-of-the-art bidirectional LSTMs \cite{sak2014long} in accuracy and meanwhile the training process may be accelerated by more than 3 times. Furthermore, the proposed FSMNs introduce much smaller latency than the bidirectional LSTMs, making it suitable for many real-time applications. 

The rest of this paper is organized as follows. In section \ref{sec.1}, we introduce the architecture of the proposed FSMN model and compare it with the conventional RNNs.  In section \ref{sec.2}, we present the learning algorithm for FSMNs and an efficient implementation on GPUs. Experimental results on speech recognition  and language modelling are given and discussed in section \ref{sec.3}. Finally, the paper is concluded with our findings and future work.

\section{Feedforward Sequential Memory Networks}
\label{sec.1}

In this section, we will introduce the architecture of \emph{feedforward sequential memory networks} (FSMN), see \cite{zhang2015FSMN} for an earlier short description.  

\subsection{Model Description of FSMNs}

\begin{figure}
	\centering
	\includegraphics[width=1.0\linewidth]{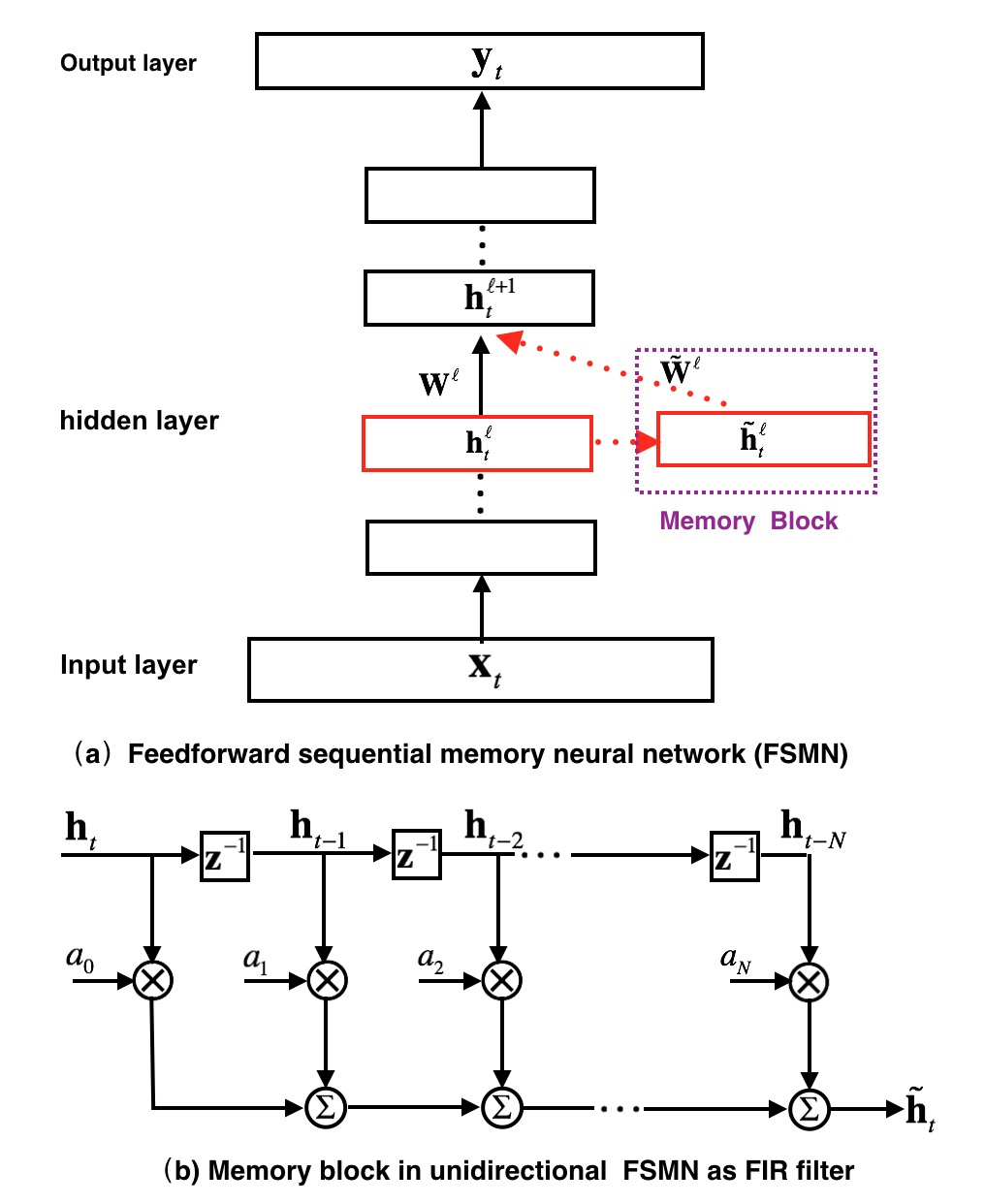}
	\caption{Illustration of a feedforward sequential memory network (FSMN) and its tapped-delay memory block. 
	(Each $z^{-1}$ block stands for a delay or memory unit)
	}
	\label{fig:FSMN_FIR}
\end{figure}

The FSMN is essentially a standard feedforward fully connected neural network with some memory blocks appeneded to the hidden layers. For instance,  Figure \ref{fig:FSMN_FIR} (a) shows an FSMN with one memory block added into its $\ell$-th hidden layer.  The memory block, as shown in Figure \ref{fig:FSMN_FIR} (b), is used to encode $N$ previous activities of the hidden layer into a fixed-size representation (called an $N$-th order FSMN), which is fed into the next hidden layer along with the current hidden activity. Depending on the encoding method to be used,  we have proposed two different variants: i) scalar FSMNs using scalar encoding coefficients ({\em sFSMN} for short); ii) vectorized FSMNs using vector encoding coefficients ({\em vFSMN} for short).

Given an input sequence, denoted as ${\bf X} = \{{\bf x}_1, \cdots,  {\bf x}_T \}$,  where each ${\bf x}_t \in \mathcal{R}^{D\times 1}$ represents the input data at time instance $t$. We further denote the corresponding outputs of the $\ell$-th hidden layer for the whole sequence as ${\bf H}^{\ell} = \{{\bf h}^{\ell}_1,  \cdots,  {\bf h}^{\ell}_T \}$, with ${\bf h}^{\ell}_t \in \mathcal{R}^{D_\ell\times 1}$. 
For an $N$-th order scalar FSMN, at each time instant $t$, we use a set of $N+1$ scalar coefficients, $\{ a_i^{\ell} \}$, to encode ${\bf h}^{\ell}_t$ and its previous $N$ terms at the $\ell$-th hidden layer into a fixed-sized representation, ${\bf{\tilde h}}^{\ell}_t$, as the output from the memory block at time $t$:
\begin{equation}\label{eq.1}
   {\bf{\tilde h}}^{\ell}_t =  \sum\limits_{i = 0}^{N} a_i^{\ell}  \cdot {\bf h}^{\ell}_{t - i} 
\end{equation}
where ${\bf a}^{\ell}=\{a^{\ell}_0, a^{\ell}_1,\cdots, a^{\ell}_{N}\}$ denote all $N+1$ time-invariant coefficients.
It is possible to use other nonlinear encoding functions for the memory blocks. In this work, we only consider linear functions for simplicity.  

As for the vectorized FSMN (vFSMN), we instead use a group of $N+1$ vectors to encode the history as follows:
\begin{equation}\label{eq.2}
   {\bf{\tilde h}}^{\ell}_t =  \sum\limits_{i = 0}^{N} {\bf a}^{\ell}_i  \odot {\bf h}^{\ell}_{t - i}
\end{equation}   
where $\odot$ denotes element-wise multiplication of two equally-sized vectors and  all coefficient vectors are denoted as: ${\bf A}^{\ell}=\{ {\bf a}^{\ell}_0, {\bf a}^{\ell}_1, \cdots, {\bf a}^{\ell}_{N} \}$.  

Obviously,  all hidden nodes share the same group of encoding coefficients in a scalar FSMN while a vectorized FSMN adopts different encoding coefficients for different hidden nodes, which may significantly improve the model capacity.  
However, a scalar FSMN has the advantage that it only introduces very few new parameters to the model and thus it can be expanded to a very high order almost without any extra cost. 

In the above FSMN definitions in eq.(\ref{eq.1}) and eq.(\ref{eq.2}), we call them unidirectional FSMNs since we only consider the past information in a sequence. These unidirectional FSMNs are suitable for some applications where only the past information is available, such as language modeling. However, in many other applications, it is possible to integrate both the history information in the past as well as certain future information within a look-ahead window from the current location of the sequence. Therefore, we may extend the above unidirectional FSMNs to the following bidirectional versions:
\begin{equation}\label{eq.bi_scalar}
   {\bf{\tilde h}}^{\ell}_t = \sum\limits_{i = 0}^{N_1} a_i^{\ell}  \cdot {\bf h}^{\ell}_{t - i} + \sum\limits_{j = 1}^{N_2}{c_j}^{\ell} \cdot {\bf h}^{\ell}_{t + j}  
\end{equation}
 \begin{equation}\label{eq.bi_vector}
   {\bf{\tilde h}}^{\ell}_t = \sum\limits_{i = 0}^{N_1} {\bf a}_i^{\ell}  \odot {\bf h}^{\ell}_{t - i} + 
   \sum\limits_{j = 1}^{N_2} {\bf c}_j^{\ell} \odot {\bf h}^{\ell}_{t + j}  
 \end{equation}
where $N_1$ is called the lookback order,  denoting the number of historical items looking back to the past, and $N_2$ the lookahead order, representing the size of the look-ahead window into the future. \footnote{In eqs.(\ref{eq.1}) to (\ref{eq.bi_vector}), for notational simplicity, we simply assume zero-padded vectors are used whenever the subscript index is out of range.}
 
The output from the memory block, ${\bf{\tilde h}}^{\ell}_t$, may be regarded as a fixed-size representation of the long surrounding  context at time instance $t$. As shown in Figure \ref{fig:FSMN_FIR} (a), ${\bf{\tilde h}}^{\ell}_t$ can be fed into the next hidden layer in the same way as ${\bf h}^{\ell}_t$.  As a result, we can calculate the activation of the units in the next hidden layer as follows:
\begin{equation}\label{eq.3}
    {\bf h}_t^{\ell  + 1} = f({\bf W}^\ell  \mathbf{h}_t^\ell  + {{{\mathbf{\tilde W}}}^\ell } {\mathbf{\tilde h}}_t^\ell  + {{\bf b}^\ell })
\end{equation}
where ${\bf W}^\ell$ and ${\bf b}^\ell$ represent the standard weight matrix and bias vector for layer $\ell$, and ${{{\mathbf{\tilde W}}}^\ell }$ denotes the weight matrix between the memory block and the next layer.



\subsection{Analysis of FSMNs}

\begin{figure}
	\centering
	\includegraphics[width=1.0\linewidth,height=4.5cm]{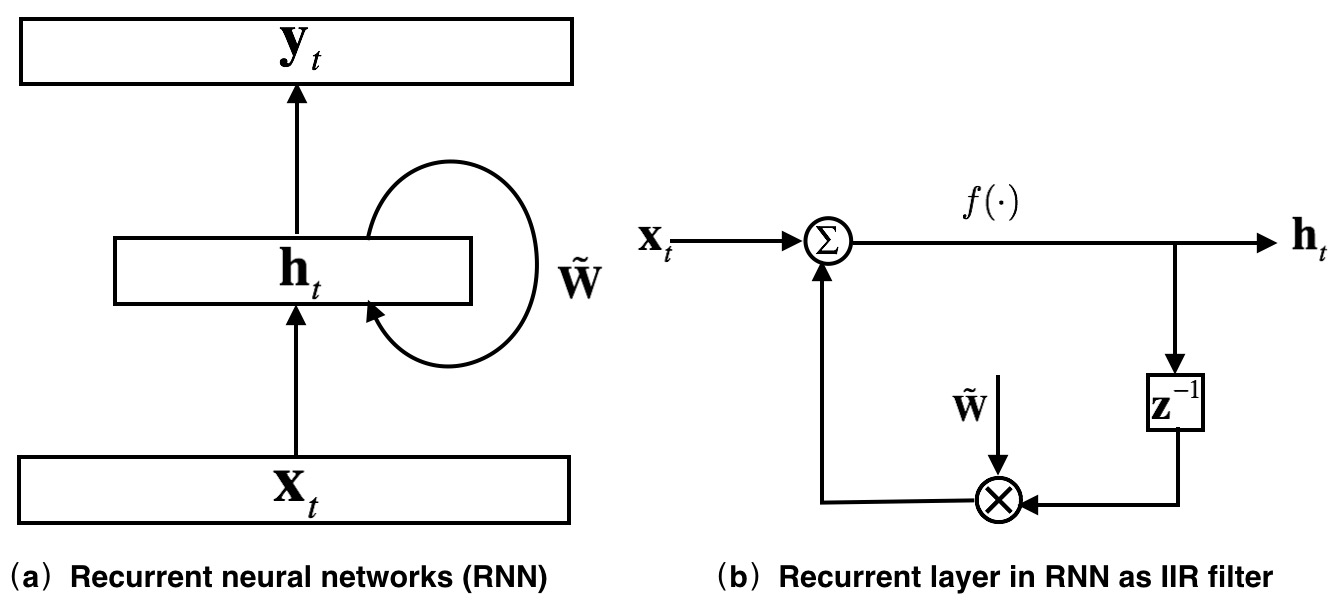}
	\caption{Illustration of recurrent neural networks and IIR-filter-like recurrent layer.}
	\label{fig:RNN}
\end{figure}

Here we analyse the properties of FSMNs and RNNs from the viewpoint of filtering in digital signal processing. Firstly, let us choose the simple recurrent neural networks (RNN) \cite{elman1990finding} as example. As shown in Figure \ref{fig:RNN} (a),  it adopts a time-delayed feedback in the hidden layer to recursively encode the history into a fixed-size representation to capture the long term dependency in a sequence. The directed cycle 
allows RNNs to exhibit some dynamic temporal behaviours. Obviously, the activations of the recurrent layer in RNNs can be denoted as follows:

\begin{equation}\label{eq.4}
{\bf h}_t=f({\bf W} {\bf x}_t  +{\bf{\tilde W}}  {\bf h}_{t-1} + {\bf b}) .
\end{equation}

Secondly, as for FSMN, we choose the unidirectional scale FSMN in eq.(\ref{eq.1})  as example. An FSMN uses a group of learnable coefficients to encode the past context within a lookback window into a fixed-size representation. The resultant representation is computed as a weighted sum of the hidden activations of all previous $N$ time instances, shown as a tapped-delay structure in Figure \ref{fig:FSMN_FIR} (b). 

From the viewpoint of signal processing,  each memory block in FSMNs may be viewed as an $N$-th order finite impulse response (FIR) filter.
Similarly, each recurrent layer in RNNs may be roughly regarded as a first-order infinite impulse response (IIR) filter, as in Figure \ref{fig:RNN} (b). It is well-known that IIR filters are more compact than FIR filters but IIR filters may be difficult to implement. In some cases, IIR filters may become unstable while FIR filters are always stable.  The learning of the IIR-like RNNs is difficult since it requires to use the so-called back-propagation through time (BPTT), which significantly increases the computational complexity and may also cause the notorious problem of gradient vanishing and exploding \cite{bengio1994learning}. However, the proposed FIR-like FSMN is an overall feedforward structure that can be efficiently learned using the standard back-propagation (BP) with stochastic gradient descent (SGD).  As a result, the learning of FSMNs may be more stable and easier than that of RNNs.  
More importantly,  it is well known that any IIR filter can be approximated by a high-order FIR filter up to sufficient precision \cite{oppenheim1989discrete}. In spite of the nonlinearity of RNNs in eq.(\ref{eq.4}),  we believe FSMNs provide a good  alternative to capture the long-term dependency in sequential signals. If we set proper orders, FSMNs may work equally well as RNNs or perhaps even better.
\subsection{Attention-based FSMN}
For sFSMN and vFSMN we use context-independent coefficients to encode the long surrounding context into a fixed-size representation. In this work, we also try to use context-dependent coefficients, which we called attention-based FSMN.  We use the 
following attention function \cite{bahdanau2014neural} to calculate the context-dependent coefficients:
\begin{equation}
{\mathbf{a}}_t^\ell  = {\mathbf{V}^\ell} \cdot f({\mathbf{U}^\ell} {\mathbf{h}}_t^\ell  + {{\mathbf{m}}^\ell })
\end{equation}
where, ${\mathbf{a}}_t^\ell  \in {\mathcal{R}^{({N_1} + {N_2}) \times 1}}$ and and ${\bf V}^\ell$, ${\bf U}^\ell$, ${\bf m}^\ell$ are the parameters of the attention function. $N_1$ and $N_2$ denote the lookback and lookahead orders respectively.  As a result, ${\mathbf{a}}_t^\ell$ is a group of context-dependent coefficients with respect to $\mathbf{h}_t^\ell$, which are used to encode the long surrounding context at time instance $t$ as follow:
\begin{equation}
{\mathbf{\tilde h}}_t^\ell  = \sum\limits_{i = 0}^{{N_1} - 1} {a_{t,i}^\ell }  \cdot {\mathbf{h}}_{t - i}^\ell  + \sum\limits_{j = 1}^{{N_2}} {a_{t,{N_1} - 1 + j}^\ell }  \cdot {\mathbf{h}}_{t + j}^\ell 
\end{equation}
The same to sFSMN and vFSMN, $\mathbf{\tilde h}_t^\ell$ is fed into the next hidden layer.
\section{Efficient Learning of FSMNs}
\label{sec.2}

Here we consider how to learn FSMNs using mini-batch based stochastic gradient descent (SGD).
In the following,  we present an efficient implementation and show that the entire learning algorithm can be formulated as matrix multiplications suitable for GPUs.


For the scalar FSMNs in eq. (\ref{eq.1}) and eq. (\ref{eq.bi_scalar}), each output from the memory block is a sum of the hidden activations weighted by a group of coefficients to be learned. 
We first demonstrate the forward pass of FSMNs can be conducted as some sequence-by-sequence matrix multiplications.
Take a unidirectional $N$-th order scalar FSMN in eq. (\ref{eq.1}) as example, all $N+1$ coefficients in the memory block are assumed to be $\{ a_0, a_1,\cdots, a_N\}$,  given an input sequence $\bf{X}$ consisting of $T$ instances, we may construct a $T \times T$ upper band matrix ${\bf M}$ as follows: 
\begin{equation}\label{eq_matrix_one_scalar}
	\bf{M} = \left[ \begin{gathered}
		{a_0} \cdots {a_N}\quad 0\; \cdots \cdots \;0   \hfill \\
		0\quad {a_0} \cdots \;{a_N}\; \cdots \cdots \; 0   \hfill \\
		\vdots \;\; \cdots \; \quad  \ddots \quad \quad  \ddots \;\; \vdots  \hfill \\
		0\quad  \cdots \quad \quad {a_0}\; \cdots \;\;{a_N} \hfill \\
		\vdots \qquad  \cdots \quad \quad \quad \ddots \;\; \vdots  \hfill \\
		0\quad  \cdots \qquad \;\; 0 \quad \quad\; {a_0} \hfill \\ 
	\end{gathered}  \right].
\end{equation}
As for the bidirectional scalar FSMNs in eq. (\ref{eq.bi_scalar}), we can construct the following $T \times T$ band matrix ${\bf M}$:
\begin{equation}\label{eq_matrix_two_scalar}
	\bf{M} = \left[ \begin{gathered}
  {a_0}\;\; \cdots \;\;{a_{{N_1}}}\;\;\;0\qquad  \cdots \qquad 0 \hfill \\
  {c_1}\;\;\;{a_0}\;\; \cdots \;\;{a_{{N_1}}}\;\;\;\;0\;\;\;\; \cdots \;\;\;\; \vdots \; \hfill \\
   \vdots \quad  \ddots \;\; \ddots \qquad \;  \ddots \quad  \ddots \qquad  \vdots  \hfill \\
  {c_{{N_2}}}\; \cdots \;\;{c_1}\;\;\;{a_0}\;\; \cdots {a_{{N_1}}}\; \cdots \;0 \hfill \\
  0\quad \quad \;\qquad \; \ddots \;\; \ddots \quad \quad \quad \;\;\; \vdots  \hfill \\
   \vdots \quad \qquad {c_{{N_2}}}\qquad \;\; {c_1}\quad {a_0}\qquad {a_{{N_1}}} \hfill \\
   \vdots \qquad \qquad \; \ddots \;\;\qquad \quad \;\ddots \quad \; \vdots  \hfill \\
  0\quad  \cdots \;\;\;\;\;\;\;0\;\;\;\;{c_{{N_2}}}\;\; \cdots \;\;{c_1}\;\;{a_0} \hfill \\ 
\end{gathered}  \right].
\end{equation}
Obviously, the sequential memory operations in eq.(\ref{eq.1}) and eq.(\ref{eq.bi_scalar}) for the whole sequence can be computed as one matrix multiplication as follows: 
\begin{equation}
{\bf{\tilde H}} = {\bf H} \; {\bf M}
\end{equation}
where the matrix ${\bf H}$ is composed of all hidden activations of the whole sequence \footnote{Obviously, ${\bf H}$ can also be computed altogether in parallel for the whole sequence.}, and ${\bf{\tilde H}}$ is the corresponding outputs from the memory block for the entire sequence. Furthermore, we can easily extend the above formula to a mini-batch  consisting of $K$ sequences, i.e., ${\cal L}=\{ {\bf X}_1 \;  \cdots {\bf X}_K\}$. In this case, we can compute the memory outputs for all $K$ sequences in the mini-batch as follows:
\begin{equation}
		{\bf \tilde{H}} =\left[{\bf H}_1, {\bf H}_2 \cdots {\bf H}_K\right]  \left[ 
		\begin{gathered}
			{\bf M}_1 \hfill \\
			\qquad  {\bf M}_2 \hfill \\
			\qquad \qquad  \ddots  \hfill \\
			\qquad \qquad \qquad  {\bf M}_K\hfill \\ 
		\end{gathered}  \right] 
		= {\bf \bar{H}} \; {\bf \bar{M}}	
\end{equation}
where each ${\bf M}_k$ is constructed in the same way as eq.(\ref{eq_matrix_one_scalar}) or (\ref{eq_matrix_two_scalar}) based on the length of each sequence.

During the backward procedure, except the regular weights in the neural nets, we also need to calculate the gradients with respect to ${\bf \bar{M}}$, to update the filter coefficients. Since FSMNs remain as a pure feedforward network structure, we can calculate the gradients using the standard back-propagation (BP) algorithm. Denote the error signal with respect to $\bf {\tilde{H}}$ as  ${\bf e}_{\bf{\tilde H}}$, which is back-propagated from the upper layers, the gradients with respect to ${\bf \bar{M}}$ can be easily derived as:
\begin{equation}
  \Delta {\bf{\bar M}} = {\bf \bar{H}}^\top \; {\bf e} _{\bf{\tilde H}} .
\end{equation}
Furthermore, the error signal w. r. t. ${\bf{\bar H}}$ is computed as:
\begin{equation}
  {\bf e}_{\bf{\bar H}} = {\bf e}_{\mathbf{\tilde H}} \; {\mathbf{\bar M}}^\top .
\end{equation}
This error signal is further back-propagated downstream to the lower layers. 
As shown above, all computations in a scalar FSMN can be formulated as matrix multiplications, which can be efficiently conducted in GPUs. As a result, scalar FSMNs have low computational complexity in training, comparable with the standard DNNs.

Similarly, for unidirectional and bidirectional vectorized FSMNs, we can calculate the outputs from the memory block as a weighted sum of the hidden layer's activations using  eqs. (\ref{eq.2}) and (\ref{eq.bi_vector}), respectively.  Therefore, for the unidirectional vectorized FSMNs, we can calculate the gradients with respect to the encoding coefficients ${\bf a}^{\ell}_i$ as well as the error signals with respect to the hidden activation ${\bf h}^{\ell}_t$ as the following forms:
\begin{equation}\label{eq.10}
     \Delta {\bf a}_i^\ell  = \sum\limits_{t = 1}^T {{{\bf e} _{{\mathbf{\tilde h}}_t^\ell }} \odot {\mathbf{h}}_{t - i}^\ell } 
\end{equation}
\begin{equation}
    {{\bf e} _{{\mathbf{h}}_t^\ell }} = \sum\limits_{i = 0}^N {{\mathbf{a}}_i^\ell  \odot {{\bf e}_{{\mathbf{\tilde h}}_{t + i}^\ell }}}.
 \end{equation}
And for the bidirectional vFSMNs, the corresponding gradient and error signals are computed as follows:
\begin{equation}
 \Delta {{\mathbf{a}}_i^\ell } = \sum\limits_{t = 1}^T {{{\bf e} _{{\mathbf{\tilde h}}_t^\ell }} \odot {\mathbf{h}}_{t - i}^\ell } \;, \quad {\Delta {{\mathbf{c}}_i^\ell }} = \sum\limits_{t = 1}^T {{{\bf e} _{{\mathbf{\tilde h}}_t^\ell }} \odot {\mathbf{h}}_{t + i}^\ell } 
\end{equation}
\begin{equation}\label{eq.13}
{{\bf e}_{{\mathbf{h}}_t^\ell }} = \sum\limits_{i = 0}^{{N_1}} {{\mathbf{a}}_i^\ell  \odot {e _{{\mathbf{\tilde h}}_{t + i}^\ell }}}  + \sum\limits_{i = 1}^{{N_2}} {{\mathbf{c}}_i^\ell  \odot {e _{{\mathbf{\tilde h}}_{t - i}^\ell }}} 
\end{equation}
where ${\bf e} _{{\bf{\tilde h}}_{t}^\ell}$ is the error signal with respect to ${\bf{\tilde h}}_{t}^\ell$.
Note that these can also be computed efficiently on GPUs using CUDA kernel functions with element-wise multiplications and additions.

\section{Experiments}
\label{sec.3}

In this section, we evaluate the effectiveness and efficiency of the proposed FSMNs on several standard benchmark tasks in speech recognition and language modelling and compare with the popular LSTMs in terms of modeling performance and learning efficiency.  

\subsection{Speech Recognition}

For the speech recognition task, we use the popular Switchboard (SWB) dataset. The training data set consists of 309-hour Switchboard-I training data and 20-hour Call Home English data. We divide the whole training data into two sets: training set and cross validation set. The training set contains 99.5\% of training data, and the cross validation set contains the other 0.5\%. Evaluation is performed in terms of word error rate (WER) on the Switchboard part of the standard NIST 2000 Hub5 evaluation set (containing 1831 utterances), denoted as Hub5e00. 

\subsubsection{Baseline Systems}

For the baseline GMM-HMMs system, we train a standard tied-state cross-word tri-phone system using the 39-dimension PLPs (static, first and second derivatives) as input features. The baseline is estimated with the maximum likelihood estimation (MLE) and then discriminatively trained based on the minimum phone error (MPE) criterion.  Before the model training, all PLP features are pre-processed with the cepstral mean and variance normalization (CMVN) per conversation side. The final hidden Markov model (HMM) consists of 8,991 tied states and 40 Gaussian components per state. In the decoding, we use a trigram language model (LM) that is trained on 3 million words from the training transcripts and another 11 million  words of the Fisher English Part 1 transcripts.  The performance of the baseline MLE and MPE trained GMM-HMMs systems are 28.7\% and 24.7\% in WER respectively.

As for the DNN-HMM baseline system, we follow the same training procedure as described in \cite{dahl2012context,zhang2015rectified} to train the conventional context dependent DNN-HMMs using the tied-state alignment obtained from the above MLE trained GMM-HMMs baseline system. We have trained standard feedforward fully connected neural networks (DNN) using either sigmoid or ReLU activation functions. The DNN contains 6 hidden layers with 2,048 units per layer. The input to the DNN is the 123-dimensional log filter-bank (FBK) features concatenated from all consecutive frames within a long context window of (5+1+5). The sigmoid DNN system is first pre-trained using the RBM-based layer-wise pre-training while the ReLU DNN is randomly initialized. In the fine-tuning, we use the mini-batch SGD algorithm to optimize the frame-level cross-entropy (CE) criterion. 
The performance of baseline DNN-HMMs systems is listed in Table \ref{tab:SWB_3} (denoted as DNN-1 and DNN-2).

Recently, the hybrid long short term memory (LSTM) recurrent neural networks and hidden Markov models (LSTM-HMM) are applied to acoustic modeling \cite{abdel2012applying, sainath2013deep,sak2014long} and they have achieved the state-of-the-art performance for large scale speech recognition.  In \cite{ sainath2013deep,sak2014long}, it also introduced a projected LSTM-RNN architecture, where each LSTM layer is followed by a low-rank linear recurrent projection layer that helps to reduce the model parameters as well as accelerate the training speed. 
In this experiment, we rebuild the deep LSTM-HMM baseline systems by following the same configurations introduced in \cite{sak2014long}.  The baseline LSTM-HMM contains three LSTM layers with 2048 memory cells per layer and each LSTM layer followed by a low-rank linear recurrent projection layer of 512 units. Each input to the LSTM is 123-dimensional FBK features calculated from a 25ms speech segment. Since the information from the future frames is helpful for making a better decision for the current frame, we delay the output state label by 5 frames (equivalent to using a look-ahead window of 5 frames).  The model is trained with the truncated BPTT algorithm \cite{werbos1990backpropagation} with a time step of 16 and a mini-batch size of 64 sequences.

Moreover, we have also trained a deep bidirectional LSTM-HMMs baseline system. Bidirectional LSTM (BLSTM) can operate on each input sequence from both directions, one LSTM for the forward direction and the other for the backward direction.  As a result, it can take both the past and future information into account to make a decision for each time instance. In our work, we have trained a deep BLSTM consisting of three hidden layers and 2048 memory cells per layer (1024 for forward layer and 1024 for backward layer). Similar to the unidirectional LSTM, each BLSTM layer is also followed by a low-rank linear recurrent projection layer of 512 units. The model is trained using the standard BPTT with a mini-batch of 16 sequences. 

The performance of the LSTM and BLSTM models is listed in the fourth and fifth rows of  Table \ref{tab:SWB_3}  respectively (denoted as LSTM and BLSTM). Using BLSTM, we can achieve a low word error rate of 13.5\% in the test set. This is a very strong baseline in this task.\footnote{The previously reported best results (in WER) in the Switchboard task under the same training condition include: 15.6\% in \cite{su2013error} using a large DNN model
with 7 hidden layers plus data re-alignement; 13.5\% in \cite{saon2014unfolded} using a deep unfolded RNN with front-end feature adaptation; and 14.8\% in \cite{chen2015training} using a Bidirectional LSTM.}

\subsubsection{FSMN Results}

In speech recognition, it is better to take bidirectional information into account to make a decision for current frame. Therefore, we use the bidirectional FSMNs in eq. (\ref{eq.bi_scalar}) and eq. (\ref{eq.bi_vector}) for this task. Firstly, we have trained a scalar FSMN with 6 hidden layer and 2048 units per layer. The hidden units adopt the rectified linear (ReLU) activation function.  
The input to FSMNs is the 123-dimensional FBK features concatenated from three consecutive frames within a context window of (1+1+1). Different from DNNs, which need to use a long sliding window of acoustic frames as input, FSMNs do not need to concatenate too many consecutive frames due to its inherent memory mechanism.  In our work, we have found that it is enough to just concatenate three consecutive frames as input. The learning schedule of FSMNs is the same as the baseline DNNs.

\begin{table}
	\centering
	\caption{Performance comparison (in WER) of vectorized FSMNs (vFSMN) with various lookback orders ($N_1$) and lookahead orders ($N_2$) in the Switchboard task.}
	\begin{tabular}[t]{|c|c|c|}
		\hline
		$N_1$ &  $N_2$ & WER(\%)\\\hline
		20 & 10 & 13.7\\\hline
		20 & 20 & 13.6\\\hline
		40 & 40 & 13.4\\\hline
		50 & 50 & {\bf 13.2} \\\hline
		100 & 100 & 13.3\\\hline
	\end{tabular}
	\label{tab:fsmn_dif_order}
\end{table}

In the first experiment, we have investigated the influence of the various lookback and lookahead orders of bidirectional FSMNs on the final speech recognition performance.  We have trained several vectorized FSMNs with various lookback and lookahead order configurations. Experimental results are shown in Table \ref{tab:fsmn_dif_order}, from which we can see that vFSMN can achieve a WER of 13.2\% when the lookback and lookahead orders are both set to be 50.  To our best knowledge, this is the best performance reported on this task for speaker-independent training (no speaker-specific adaptation and normalization) using the frame-level cross entropy
error criterion. In real-time speech recognition applications, we need to consider the latency. In these cases, the bidirectional LSTMs are not suitable since the backward pass can not start until the full sequence is received, which normally cause an unacceptable time delay. However, the latency of bidirectional FSMNs can be easily adjusted by reducing the lookahead order. For instance, we can still achieve a very competitive performance (13.7\% in WER) when setting the lookahead order to  10. In this case, the total latency per sequence is normally tolerable in real-time speech recognition tasks. Therefore, FSMNs are better suited for low-latency speech recognition than bidirectional LSTMs.

\subsubsection{Model Comparison}

In Table \ref{tab:SWB_3}, we have summarized experimental results of various systems on the SWB task. Results have shown that those models utilizing the long-term dependency of speech signals, such as LSTMs and FSMNs, perform much better than others. Among them, the bidirectional LSTM can significantly outperform the unidirectional LSTM since it can take the future context into account. More importantly, the proposed vectorized FSMN can slightly outperform  BLSTM, being simpler in model structure and faster in learning speed. For one epoch of learning, BLSTMs take about 22.6 hours while the vFSMN only need  about 7.1 hours, over 3 times speedup in training. 

\begin{table}
\centering
\caption{Comparison (training time per epoch in hour, recognition performance in WER) of various acoustic models in the Switchboard task. DNN-1 and DNN-2 denote the standard 6 layers of fully connected neural networks using sigmoid and ReLU activation functions. FSMN and vFSMN denote the scalar FSMN and vectorized FSMN respectively.}
\begin{tabular}[t]{|c|c|c|c|}
        \hline
	model   & time (hr)   & WER(\%) \\\hline
	DNN-1      &   5.0        & 15.6 \\\hline
	DNN-2     &   4.8        & 14.6 \\\hline
	LSTM      &   9.4         & 14.2 \\\hline
	BLSTM     &   22.6       & {\bf 13.5} \\\hline \hline
	sFSMN     &   6.7         & 14.2 \\\hline
	{\bf vFSMN}  &   7.1          & {\bf 13.2} \\\hline
	\end{tabular}
	\label{tab:SWB_3}
\end{table}

\begin{figure}
	\centering
	\includegraphics[width=1.0\linewidth]{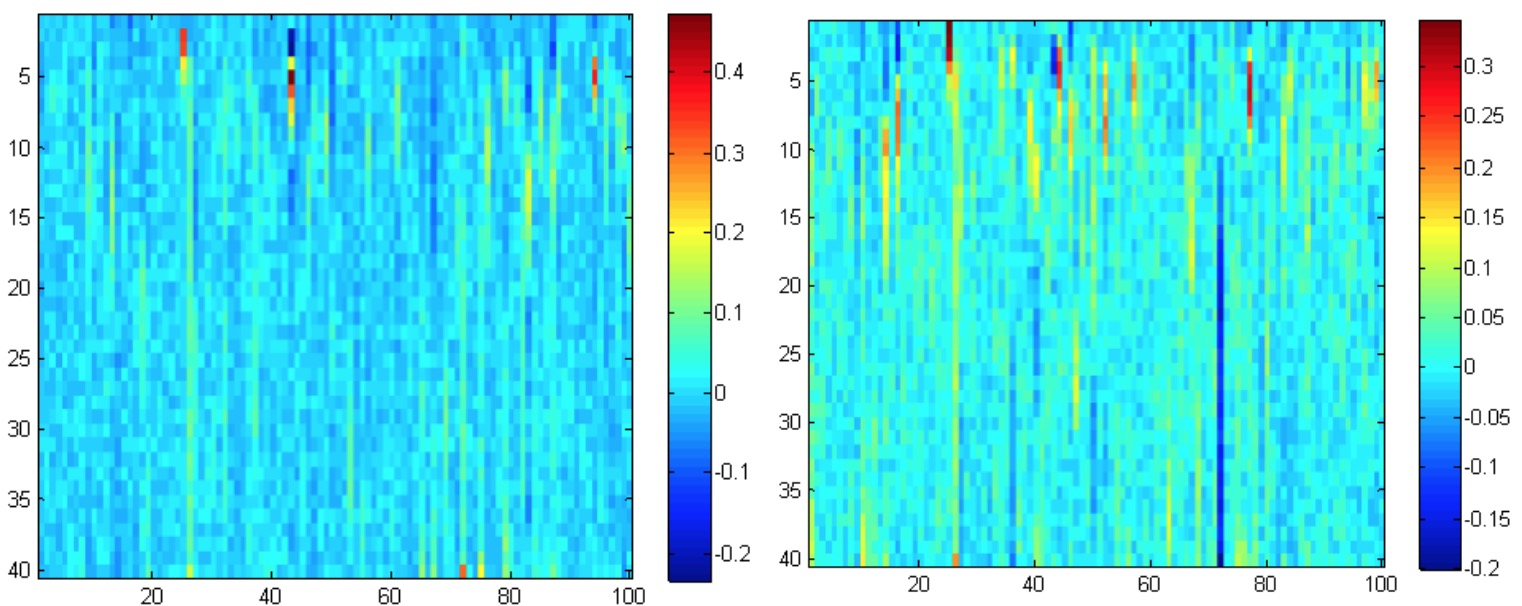}
	\caption{An illustration of the first 100 dimensions of the learned lookback filters (left) and lookahead filters (right) in a vectorized FSMN. Both the lookback  and lookahead filters are set to be 40th order.}
	\label{fig:SWB_filter}
\end{figure}

Moreover, experimental results in the last two lines of Table \ref{tab:SWB_3} also show that the vectorized FSMN perform much better than the scalar FSMN. We have investigated these results by virtualizing the learned coefficient vectors in the vectorized FSMN. In Figure \ref{fig:SWB_filter}, we have shown the learned filter vectors in the first memory layer in the vFSMN model.  We can see that different dimensions in the memory block have learned quite  different filters for speech signals. 
As a result, vectorized FSMNs perform much better than the scalar FSMNs for speech recognition.

\subsubsection{Attention-based FSMN}
In this section, we will compare the performance of attention-based FSMN with DNN and vFSMN.  we used the 39 dimension PLP feature as inputs. All models contain 6 hidden layers with 2048 units per layer and use ReLU as the activation functions. The lookback and lookahead orders are 40 for both attention-based FSMN and vFSMN. Experimental results are shown in Table \ref{tab:SWB_4}.  Attention-based FSMN can achieve a significant improvement in FACC (67.42\% to 48.64\%). However, the improvement of the word error rate (WER) is small over the DNN baseline (15.3\% to 15.6\%). Moreover, this experiment shows that the attention-based FSMN performs significantly worse than the regular vFSMN without using the attention mechanism. 
\begin{table}
	\centering
	\caption{ Performance (Frame classification accuracy in FACC, recognition performance in WER) of the attention-based FSMN model in the Switchboard task. }
\begin{tabular}[t]{|c|c|c|}
	\hline
	model   & FACC(\%)  & WER(\%) \\\hline
	RL-DNN      &   48.64        & 15.6 \\\hline
	{\bf vFSMN}  &   67.42          & {\bf 13.8} \\\hline
	Attention-FSMN & 65.16  & 15.3 \\\hline
\end{tabular}
\label{tab:SWB_4}
\end{table}


\subsection{Language Modeling}

A statistical language model (LM) is a probability distribution over sequences of words.  Recently, neural networks have been successfully applied to language modeling \cite{bengio2003neural,mikolov2010recurrent}, yielding the state-of-the-art performance. The basic idea of neural network language models is to use a projection layer to map discrete words into a continuous space and estimate word conditional probabilities in this space, which may be smoother to better generalized to unseen contexts.  In language modeling tasks, it is quite important to take advantage of the long-term dependency of a language. Therefore, it is widely reported that RNN based LMs can outperform FNN based LMs in language modeling tasks. The so-called FOFE \cite{zhang2015fixed} based method provides another choice to model long-term dependency for languages.  

Since the goal in language modeling is to predict next word in a text sequence given all previous words. Therefore, different from speech recognition, we can only use the unidirectional FSMNs in eq.(\ref{eq.1}) and eq.(\ref{eq.2}) to evaluate their capacity in learning long-term dependency of language.  We have evaluated the FSMN based language models (FSMN-LMs) on two tasks: i) the Penn Treebank (PTB) corpus of about 1M words. The vocabulary size is limited to 10k. The preprocessing method and the way to split data into training/validation/test sets are the same as \cite{mikolov2011extensions}. 
 ii) The English wiki9 dataset, which is composed of  the first $10^9$ bytes of English wiki data as in \cite{Mahoney2011}. We split it into three parts:  training (153M), validation (8.9M) and test (8.9M) sets.   The vocabulary size is limited to 80k for wiki9 and replace all out-of-vocabulary words by $<$UNK$>$.  Details of the two datasets can be found in Table \ref{tab:datasets}. 
 
 \begin{table}[t]
 	\centering
 	\caption{The sizes of the PTB and English wiki9 corpora are given in number of words.}
 	\begin{tabular}{|c|c|c|c|}
 		\hline  
 		Corpus & train & valid & test \\ \hline  
 		PTB & 930k & 74k &  82k \\ \hline
 		wiki9 & 153M & 8.9M & 8.9M \\ 
 		\hline 
 	\end{tabular} 
 	\label{tab:datasets}
 \end{table}

\subsubsection{Training Details} 

For the FSMNs, all hidden units adopt the rectified linear activation function.  In all experiments, the networks are randomly initialized, without using any pre-training method.  We use SGD with a mini-batch size of 200 and 500 for PTB and English wiki9 tasks respectively. The initial learning rate is set to 0.4, which is kept fixed as long as the perplexity on the validation set decreases by at least 1. After that, we continue six more epochs of training, where the learning rate is halved after each epoch. Because PTB is a very small task, we also use momentum (0.9) and weight decay (0.00004) to avoid overfitting. For the wiki9 task, we do not use the momentum or weight decay.

\subsubsection{Performance Comparison}

\begin{table}[t]
	\centering
	\caption{Perplexities on the PTB database for various LMs.}
	\begin{tabular}{|l|c|}
		\hline 
		Model  & Test PPL \\\hline 
		KN 5-gram \cite{mikolov2011extensions}      &  141 \\
	    3-gram FNN-LM \cite{zhang2015fixed} & 131 \\
		RNN-LM \cite{mikolov2011extensions}  & 123 \\
		LSTM-LM \cite{graves2013generating}   &  117   \\ 
		MemN2N-LM \cite{sukhbaatar2015end} & 111\\
		FOFE-LM \cite{zhang2015fixed} & 108 \\ 
		Deep RNN \cite{pascanu2013construct}& 107.5\\
		Sum-Prod Net \cite{cheng2014language} &100\\\hline
		LSTM-LM (1-layer)    & 114   \\ 
		LSTM-LM (2-layer)    & 105   \\   \hline
		{\bf sFSMN-LM}                                & \textbf{102}\\
		{\bf vFSMN-LM}                                 & \textbf{101}\\\hline
	\end{tabular} 
	\label{tab:PTB_summary}
\end{table}

For the PTB task, we have trained both scalar and vector based FSMNs with an input context window of two, where the previous two words are sent to the model at each time instance to predict the next word. Both models contain a linear projection layer (of 200 units), two hidden layers (of 400 units pre layer) and a memory block in the first hidden layer. We use a 20th order FIR filter in the first hidden layer for both scalar FSMNs and vectorized FSMNs in the PTB task.
These models can be trained in 10 minutes on a single GTX780 GPU. For comparison, we have also builded two LSTM based LMs with Theano \cite{bergstra2011theano}, one using one recurrent layer and the other using two recurrent layers.  In Table \ref{tab:PTB_summary}, we have summarized the perplexities on the PTB test set for various language models.\footnote{All the models in Table \ref{tab:PTB_summary} do not use the dropout regularization, which is somehow equivalent to data augmentation.  In \cite{zaremba2014recurrent,kim2015character}, the proposed LSTM-LMs (word level or character level) achieves much lower perplexity but they both use the dropout regularization and take days to train a model.} 

For the wiki9 task, we have trained several baseline systems: traditional n-gram LM,  RNN-LM, standard FNN-LM and the FOFE-LM introduced in \cite{zhang2015fixed}. Firstly, we have trained two n-gram LMs (3-gram and 5-gram) using the modified Kneser-Ney smoothing without count cutoffs. As for RNN-LMs,  we have trained a simple RNN with one hidden layer of 600 units using the toolkit in \cite{mikolov2010recurrent}. We have further used the spliced sentence method in \cite{chen2014efficient} to speed up the training of RNN-LM on GPUs. The architectures of FNN-LM and FOFE-LM are the same, it contains a linear projection layer of 200 units and three hidden layer with 600 units per layer and hidden units adopt the ReLU activation. The only difference is that FNN-LM uses the one-hot vectors as input while FOFE-LM uses the so-called FOFE codes as input. In both models, the input window size is set to two words. The performance of the baseline systems is listed in Table \ref{tab:wiki9_summary}.  For the FSMN based language models, we use the same architecture as the baseline FNN-LM.  Both the scalar and vector based FSMN adopt a 30th order FIR filter in the memory block for the wiki9 task.  In these experiments, we have also evaluated several  FSMN-LMs with memory blocks added to different hidden layers. Experimental results on the wiki9 test set are listed in Table \ref{tab:wiki9_summary} for various LMs.

\begin{table}[t]
	\centering
	\caption{Perplexities on the English wiki9 test set for various language models ($M$ denotes a hidden layer with memory block).} 
	\begin{tabular}{|l|l|c|}
		\hline
		Model & Architecture & PPL \\\hline
		KN 3-gram &   -                &    156 \\
		KN 5-gram &  -                 &   132  \\\hline
		FNN-LM     & [2*200]-3*600-80k & 155 \\\hline      
		RNN-LM   & [1*600]-80k   &   112 \\\hline          
		FOFE-LM & [2*200]-3*600-80k &  104\\\hline
		sFSMN-LM & [2*200]-600(M)-600-600-80k & 95\\
		       &  [2*200]-600-600(M)-600-80k & 96\\
		 &  [2*200]-600(M)-600(M)-600-80k & \textbf{92}\\\hline
		 vFSMN-LM &[2*200]-600(M)-600-600-80k & 95\\
		    & [2*200]-600(M)-600(M)-600-80k & \textbf{90}\\\hline
	\end{tabular} 
	\label{tab:wiki9_summary}
\end{table}

\begin{figure}
	\centering
	\includegraphics[width=0.8\linewidth]{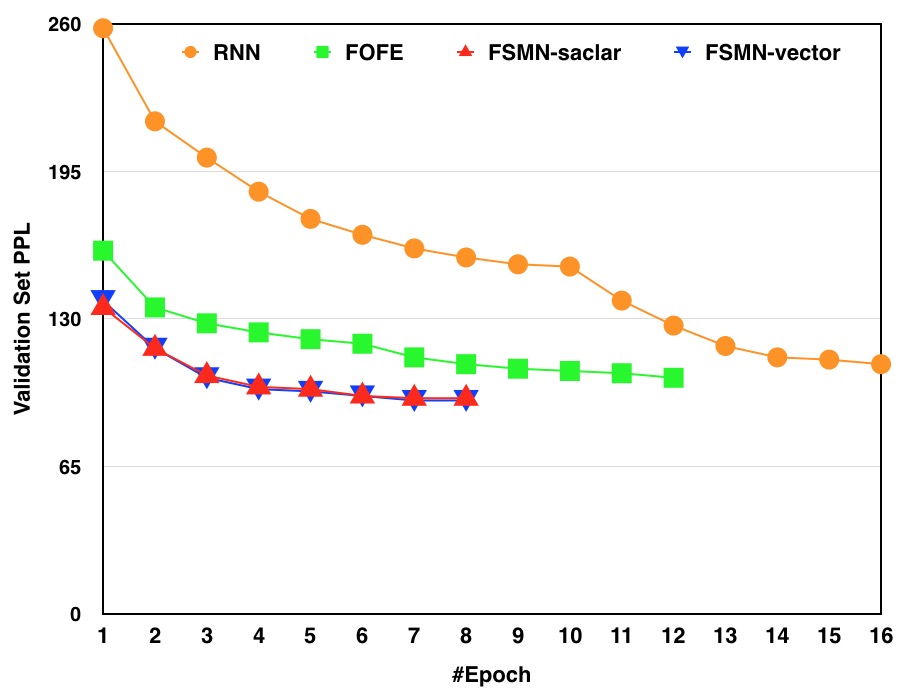}
	\caption{The learning curves of various models on the English wiki9 task.}
	\label{fig.wiki_curve}
\end{figure}

From the experimental results in Table \ref{tab:PTB_summary} and Table \ref{tab:wiki9_summary}, we can see that the proposed FSMN based LMs can significantly outperform not only the traditionally FNN-LM but also the RNN-LM and FOFE-LM.  For example, in the English wiki9 task, the proposed FSMN-LM can achieve a perplexity of 90 while the well-trained RNN-LM and FOFE-LM obtain 112 and 104 respectively. This is the state-of-the-art performance for this task. Moreover, the learning curves in Figure \ref{fig.wiki_curve} have shown that the FSMN based models converge much faster than RNNs. It only takes about 5 epochs of learning for FSMNs while RNN-LMs normally need more than 15 epochs. Therefore, training an FSMN-LM is much faster than an RNN-LM. Overall, experimental results indicate that FSMNs can effectively encode long context into a compressed fixed-size representation, being able to explore the long-term dependency in text sequences.

\begin{figure}
	\centering
	\includegraphics[width=1.0\linewidth]{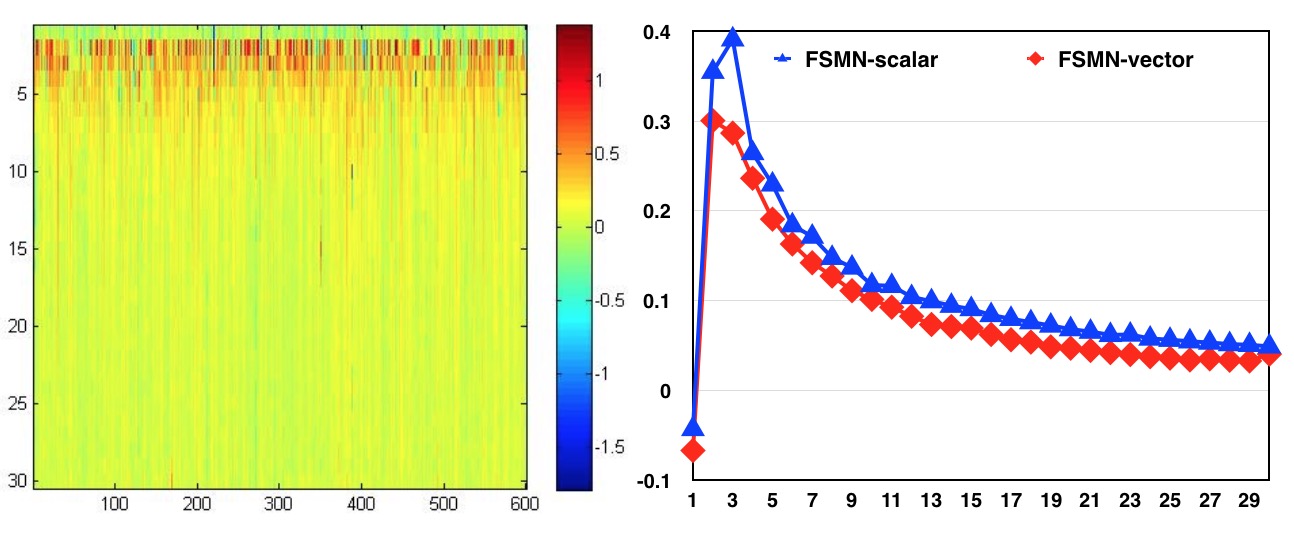}
	\caption{Illustration of the learned filters in FSMNs on the PTB task: left) the coefficients of the learned filters in vectorized FSMN; right) the average coefficients of filters in vFSMN and  the learned filters in the scalar based FSMN.}
	\label{fig:wiki_filter}
\end{figure}

Another interesting finding is that the scalar and vector based FSMN-LMs achieve similar performance on both PTB and wiki9 tasks. This is very different from the experimental results on the speech recognition task (see Table \ref{tab:SWB_3}), where vectorized FSMN model significantly outperforms the scalar FSMN models.   
Here we have investigated the learned coefficients of the FIR filter in two FSMN-LMs.
We choose the well-trained scalar and vector based FSMN models with a memory block in the first hidden layer.
In Figure \ref{fig:wiki_filter}, we have plotted the learned filter coefficients in both vector and scalar based FSMNs. The motivation to use a vectorized FSMN is to learn different filters for various data dimensions. However, from the left figure in Figure \ref{fig:wiki_filter}, we can see that the  learned filters of all dimension are very similar in the vectorized FSMN. Moreover, the averaged (across all dimensions) filter coefficients of vectorized FSMN match very well with the filter learned by the scalar FSMN, as shown in the right figure in Figure \ref{fig:wiki_filter}. This explains why the scalar and vector based FSMN-LMs achieve similar performance in language modeling tasks. Finally, we can see that the learned filter coefficients reflect the property of nature language that nearby contexts generally play more important role in prediction than far-away ones.


\section{Conclusions and Future Work}
\label{sec:conclusion-and-future-work}

In summary, we have proposed a novel neural network architecture, namely feedforward sequential memory networks (FSMN)  for modeling long-term dependency in sequential data. The memory blocks in FSMNs use a tapped-delay line structure to encode long context information into a fixed-size representation in a pure feedforward way without using the expensive recurrent feedbacks. We have evaluated the performance of FSMNs on several speech recognition and language modeling tasks. In all examined tasks, the proposed FSMN based models can significantly outperform the popular RNN or LSTM based models. More importantly, the learning of FSMNs is much easier and faster than that of RNNs or LSTMs. As a strong alternative, we expect the proposed FSMN models may replace RNN and LSTM in a variety of tasks. As the future work, we will try to use more complex encoding coefficients, such as matrix. We will also try to apply FSMNs to other machine learning tasks under the currently popular sequence-to-sequence framework, such as question answering and machine translation. Moreover, the unsupervised learning method in \cite{Shiliang2015a,Shiliang2015b} may be applied to FSMNs to conduct unsupervised learning for sequential data. 


\bibliography{mybib}

\begin{thebibliography}{48}
\providecommand{\natexlab}[1]{#1}
\providecommand{\url}[1]{\texttt{#1}}
\expandafter\ifx\csname urlstyle\endcsname\relax
  \providecommand{\doi}[1]{doi: #1}\else
  \providecommand{\doi}{doi: \begingroup \urlstyle{rm}\Url}\fi

\bibitem[Abdel-Hamid et~al.(2012)Abdel-Hamid, Mohamed, Jiang, and
  Penn]{abdel2012applying}
Abdel-Hamid, O., Mohamed, A., Jiang, H., and Penn, G.
\newblock Applying convolutional neural networks concepts to hybrid {NN-HMM}
  model for speech recognition.
\newblock In \emph{Proceedings of IEEE International Conference on Acoustics,
  Speech and Signal Processing (ICASSP)}, pp.\  4277--4280, 2012.

\bibitem[Bahdanau et~al.(2014)Bahdanau, Cho, and Bengio]{bahdanau2014neural}
Bahdanau, D., Cho, K., and Bengio, Y.
\newblock Neural machine translation by jointly learning to align and
  translate.
\newblock \emph{arXiv preprint arXiv:1409.0473}, 2014.

\bibitem[Bengio et~al.(1994)Bengio, Simard, and Frasconi]{bengio1994learning}
Bengio, Y., Simard, P., and Frasconi, P.
\newblock Learning long-term dependencies with gradient descent is difficult.
\newblock \emph{IEEE Transactions on Neural Networks}, 5\penalty0 (2):\penalty0
  157--166, 1994.

\bibitem[Bengio et~al.(2003)Bengio, Ducharme, Vincent, and
  Janvin]{bengio2003neural}
Bengio, Y., Ducharme, R., Vincent, P., and Janvin, C.
\newblock A neural probabilistic language model.
\newblock \emph{Journal of Machine Learning Research}, 3:\penalty0 1137--1155,
  2003.

\bibitem[Bergstra et~al.(2011)Bergstra, Bastien, Breuleux, and
  et~al.]{bergstra2011theano}
Bergstra, J., Bastien, F., Breuleux, O., and et~al.
\newblock Theano: Deep learning on gpus with python.
\newblock In \emph{NIPS 2011, BigLearning Workshop, Granada, Spain}, 2011.

\bibitem[Chen et~al.(2015)Chen, Yan, and Huo]{chen2015training}
Chen, K., Yan, Z.~J., and Huo, Q.
\newblock Training deep bidirectional {LSTM} acoustic model for {LVCSR} by a
  context-sensitive-chunk {BPTT} approach.
\newblock In \emph{Proceedings of Interspeech}, 2015.

\bibitem[Chen et~al.(2014)Chen, Wang, Liu, Gales, and
  Woodland]{chen2014efficient}
Chen, X., Wang, Y.~Q., Liu, X.~Y., Gales, M.~J., and Woodland, P.~C.
\newblock Efficient gpu-based training of recurrent neural network language
  models using spliced sentence bunch.
\newblock \emph{Proceedings of Interspeech}, 2014.

\bibitem[Cheng et~al.(2014)Cheng, Kok, Pham, and et~al.]{cheng2014language}
Cheng, W.C., Kok, S., Pham, H.V., and et~al.
\newblock Language modeling with sum-product networks.
\newblock In \emph{Proceedings of Interspeech}, 2014.

\bibitem[Cho et~al.(2014)Cho, Van~Merri{\"e}nboer, and
  Gulcehre]{cho2014learning}
Cho, K., Van~Merri{\"e}nboer, B., and Gulcehre, C. et~al.
\newblock Learning phrase representations using rnn encoder-decoder for
  statistical machine translation.
\newblock \emph{arXiv:1406.1078}, 2014.

\bibitem[Chung et~al.(2014)Chung, Gulcehre, and Cho]{Chung2014Empirical}
Chung, J., Gulcehre, C., and Cho, K. et~al.
\newblock Empirical evaluation of gated recurrent neural networks on sequence
  modeling.
\newblock \emph{arXiv:1412.3555}, 2014.

\bibitem[Ciresan et~al.(2011)Ciresan, Meier, Masci, L., and
  Schmidhuber]{ciresan2011flexible}
Ciresan, D.~C., Meier, U., Masci, J., L., Maria~G., and Schmidhuber, J.
\newblock Flexible, high performance convolutional neural networks for image
  classification.
\newblock In \emph{Proceedings of International Joint Conference on Artificial
  Intelligence (IJCAI)}, pp.\  1237--1242, 2011.

\bibitem[Dahl et~al.(2012)Dahl, Yu, Deng, and Acero]{dahl2012context}
Dahl, G.E., Yu, D., Deng, L., and Acero, A.
\newblock Context-dependent pre-trained deep neural networks for
  large-vocabulary speech recognition.
\newblock \emph{IEEE Transactions on Audio, Speech, and Language Processing},
  20\penalty0 (1):\penalty0 30--42, 2012.

\bibitem[Elman(1990)]{elman1990finding}
Elman, J.~L.
\newblock Finding structure in time.
\newblock \emph{Cognitive science}, 14\penalty0 (2):\penalty0 179--211, 1990.

\bibitem[Gers et~al.(2000)Gers, Schmidhuber, and Cummins]{gers2000learning}
Gers, F.~A., Schmidhuber, J., and Cummins, F.
\newblock Learning to forget: Continual prediction with lstm.
\newblock \emph{Neural computation}, 12\penalty0 (10):\penalty0 2451--2471,
  2000.

\bibitem[Graves(2013)]{graves2013generating}
Graves, A.
\newblock Generating sequences with recurrent neural networks.
\newblock \emph{arXiv:1308.0850}, 2013.

\bibitem[Graves et~al.(2013)Graves, Mohamed, and Hinton]{graves2013speech}
Graves, A., Mohamed, A., and Hinton, G.~E.
\newblock Speech recognition with deep recurrent neural networks.
\newblock In \emph{Proceedings of IEEE International Conference on Acoustics,
  Speech and Signal Processing (ICASSP)}, pp.\  6645--6649, 2013.

\bibitem[Graves et~al.(2014)Graves, Wayne, and Danihelka]{Graves2014NTM}
Graves, A., Wayne, G., and Danihelka, I.
\newblock Neural turing machines.
\newblock \emph{arXiv:1412.3555}, 2014.

\bibitem[Hochreiter \& Schmidhuber(1997)Hochreiter and
  Schmidhuber]{hochreiter1997long}
Hochreiter, S. and Schmidhuber, J.
\newblock Long short-term memory.
\newblock \emph{Neural computation}, 9\penalty0 (8):\penalty0 1735--1780, 1997.

\bibitem[Hubel \& Wiesel(1962)Hubel and Wiesel]{hubel1962receptive}
Hubel, D.~H. and Wiesel, T.~N.
\newblock Receptive fields, binocular interaction and functional architecture
  in the cat's visual cortex.
\newblock \emph{The Journal of Physiology}, 160\penalty0 (1):\penalty0 106,
  1962.

\bibitem[Jarrett et~al.(2009)Jarrett, Kavukcuoglu, Ranzato, and
  LeCun]{jarrett2009best}
Jarrett, K., Kavukcuoglu, K., Ranzato, M., and LeCun, Y.
\newblock What is the best multi-stage architecture for object recognition?
\newblock In \emph{Proceedings of International Conference on Computer Vision
  (ICCV)}, pp.\  2146--2153, 2009.

\bibitem[Kim et~al.(2015)Kim, Jernite, Sontag, and Rush]{kim2015character}
Kim, Y., Jernite, Y., Sontag, D., and Rush, A.~M.
\newblock Character-aware neural language models.
\newblock \emph{arXiv:1508.06615}, 2015.

\bibitem[Krizhevsky et~al.(2012)Krizhevsky, Sutskever, and
  Hinton]{krizhevsky2012imagenet}
Krizhevsky, A., Sutskever, I., and Hinton, G.~E.
\newblock Imagenet classification with deep convolutional neural networks.
\newblock In \emph{Proceedings of Advances in neural information processing
  systems (NIPS)}, pp.\  1097--1105, 2012.

\bibitem[Lawrence et~al.(1997)Lawrence, Giles, Tsoi, and
  Back]{lawrence1997face}
Lawrence, S., Giles, C.~L., Tsoi, A.~C., and Back, A.~D.
\newblock Face recognition: A convolutional neural-network approach.
\newblock \emph{IEEE Transactions on Neural Networks}, 8\penalty0 (1):\penalty0
  98--113, 1997.

\bibitem[LeCun et~al.(1998)LeCun, Bottou, Bengio, and
  Haffner]{lecun1998gradient}
LeCun, Y., Bottou, L., Bengio, Y., and Haffner, P.
\newblock Gradient-based learning applied to document recognition.
\newblock \emph{Proceedings of the IEEE}, 86\penalty0 (11):\penalty0
  2278--2324, 1998.

\bibitem[LeCun et~al.(2015)LeCun, Bengio, and Hinton]{lecun2015deep}
LeCun, Y., Bengio, Y., and Hinton, G.~E.
\newblock Deep learning.
\newblock \emph{Nature}, 521\penalty0 (7553):\penalty0 436--444, 2015.

\bibitem[Mahoney(2011)]{Mahoney2011}
Mahoney, M.
\newblock Large text compression benchmark.
\newblock In \emph{http://mattmahoney.net/dc/textdata.html}, 2011.

\bibitem[Mikolov et~al.(2010)Mikolov, Karafi{\'a}t, Burget, Cernock{\`y}, and
  Khudanpur]{mikolov2010recurrent}
Mikolov, T., Karafi{\'a}t, M., Burget, L., Cernock{\`y}, J., and Khudanpur, S.
\newblock Recurrent neural network based language model.
\newblock In \emph{Proceedings of Interspeech}, pp.\  1045--1048, 2010.

\bibitem[Mikolov et~al.(2011)Mikolov, Kombrink, Burget, and
  Khudanpur]{mikolov2011extensions}
Mikolov, T., Kombrink, S., Burget, L.and~{\v{C}}ernock{\`y}, J.H., and
  Khudanpur, S.
\newblock Extensions of recurrent neural network language model.
\newblock In \emph{Proceedings of IEEE International Conference on Acoustics,
  Speech and Signal Processing (ICASSP)}, pp.\  5528--5531, 2011.

\bibitem[Nair \& Hinton(2010)Nair and Hinton]{nair2010rectified}
Nair, V. and Hinton, G.~E.
\newblock Rectified linear units improve restricted boltzmann machines.
\newblock In \emph{Proceedings of International Conference on Machine Learning
  (ICML)}, pp.\  807--814, 2010.

\bibitem[Oppenheim et~al.(1989)Oppenheim, Schafer, Buck, and
  et~al.]{oppenheim1989discrete}
Oppenheim, A.~V., Schafer, R.~W., Buck, J.~R., and et~al.
\newblock \emph{Discrete-time signal processing}, volume~2.
\newblock Prentice-hall Englewood Cliffs, 1989.

\bibitem[Pascanu et~al.(2013)Pascanu, Gulcehre, Cho, and
  Bengio]{pascanu2013construct}
Pascanu, R., Gulcehre, C., Cho, K., and Bengio, Y.
\newblock How to construct deep recurrent neural networks.
\newblock \emph{arXiv:1312.6026}, 2013.

\bibitem[Sainath et~al.(2013)Sainath, Mohamed, Kingsbury, and
  Ramabhadran]{sainath2013deep}
Sainath, T.N., Mohamed, A., Kingsbury, B., and Ramabhadran, B.
\newblock Deep convolutional neural networks for {LVCSR}.
\newblock In \emph{Proceedings of IEEE International Conference on Acoustics,
  Speech and Signal Processing (ICASSP)}, pp.\  8614--8618, 2013.

\bibitem[Sak et~al.(2014)Sak, Senior, and Beaufays]{sak2014long}
Sak, H., Senior, A., and Beaufays, F.
\newblock Long short-term memory based recurrent neural network architectures
  for large vocabulary speech recognition.
\newblock \emph{arXiv:1402.1128}, 2014.

\bibitem[Saon et~al.(2014)Saon, Soltau, Emami, and Picheny]{saon2014unfolded}
Saon, G., Soltau, H., Emami, A., and Picheny, M.
\newblock Unfolded recurrent neural networks for speech recognition.
\newblock In \emph{Proceedings of Interspeech}, 2014.

\bibitem[Schmidhuber(2015)]{Juergen2015}
Schmidhuber, J.
\newblock Deep learning in neural networks: An overview.
\newblock \emph{Neural Networks}, 61:\penalty0 85--117, 2015.

\bibitem[Schuster \& Paliwal(1997)Schuster and
  Paliwal]{schuster1997bidirectional}
Schuster, M. and Paliwal, K.~K.
\newblock Bidirectional recurrent neural networks.
\newblock \emph{IEEE Transactions on Signal Processing}, 45\penalty0
  (11):\penalty0 2673--2681, 1997.

\bibitem[Siegelmann \& Sontag(1995)Siegelmann and Sontag]{Siegelmann1995}
Siegelmann, H.~T. and Sontag, E.~D.
\newblock On the computational power of neural nets.
\newblock \emph{Journal of computer and system sciences}, 50\penalty0
  (1):\penalty0 132--150, 1995.

\bibitem[Su et~al.(2013)Su, Li, Yu, and Seide]{su2013error}
Su, H., Li, G., Yu, D., and Seide, F.
\newblock Error back propagation for sequence training of context-dependent
  deep networks for conversational speech transcription.
\newblock In \emph{Proceedings of IEEE International Conference on Acoustics,
  Speech and Signal Processing (ICASSP)}, pp.\  6664--6668, 2013.

\bibitem[Sukhbaatar et~al.(2015)Sukhbaatar, Szlam, Weston, and
  Fergus]{sukhbaatar2015end}
Sukhbaatar, S., Szlam, A., Weston, J., and Fergus, R.
\newblock End-to-end memory networks.
\newblock \emph{arXiv:1503.08895}, 2015.

\bibitem[Taigman et~al.(2014)Taigman, Yang, Ranzato, and
  Wolf]{taigman2014deepface}
Taigman, Y., Yang, M., Ranzato, M., and Wolf, L.
\newblock Deepface: Closing the gap to human-level performance in face
  verification.
\newblock In \emph{Proceedings of IEEE Conference on Computer Vision and
  Pattern Recognition (CVPR)}, pp.\  1701--1708, 2014.

\bibitem[Werbos(1990)]{werbos1990backpropagation}
Werbos, P.~J.
\newblock Backpropagation through time: what it does and how to do it.
\newblock \emph{Proceedings of the IEEE}, 78\penalty0 (10):\penalty0
  1550--1560, 1990.

\bibitem[Weston et~al.(2014)Weston, Chopra, and A.]{Weston2014MN}
Weston, J., Chopra, S., and A., Bordes.
\newblock Neural turing machines.
\newblock \emph{arXiv:1410.3916}, 2014.

\bibitem[Zaremba et~al.(2014)Zaremba, Sutskever, and
  Vinyals]{zaremba2014recurrent}
Zaremba, W., Sutskever, I., and Vinyals, O.l.
\newblock Recurrent neural network regularization.
\newblock \emph{arXiv:1409.2329}, 2014.

\bibitem[Zhang \& Jiang(2015)Zhang and Jiang]{Shiliang2015a}
Zhang, S. and Jiang, H.
\newblock Hybrid orthogonal projection and estimation ({HOPE}): A new framework
  to probe and learn neural networks.
\newblock \emph{arXiv:1502.00702}, 2015.

\bibitem[Zhang et~al.(2015{\natexlab{a}})Zhang, Jiang, and Dai]{Shiliang2015b}
Zhang, S., Jiang, H., and Dai, L.
\newblock The new {HOPE} way to learn neural networks.
\newblock In \emph{Deep Learning Workshop at ICML}, 2015{\natexlab{a}}.

\bibitem[Zhang et~al.(2015{\natexlab{b}})Zhang, Jiang, Wei, and
  Dai]{zhang2015FSMN}
Zhang, S., Jiang, H., Wei, S., and Dai, L.
\newblock Feedforward sequential memory neural networks without recurrent
  feedback.
\newblock \emph{arXiv:1510.02693}, 2015{\natexlab{b}}.

\bibitem[Zhang et~al.(2015{\natexlab{c}})Zhang, Jiang, Wei, and
  Dai]{zhang2015rectified}
Zhang, S., Jiang, H., Wei, S., and Dai, L.
\newblock Rectified linear neural networks with tied-scalar regularization for
  {LVCSR}.
\newblock In \emph{Proceedings of Interspeech}, pp.\  2635--2639,
  2015{\natexlab{c}}.

\bibitem[Zhang et~al.(2015{\natexlab{d}})Zhang, Jiang, Xu, Hou, and
  Dai]{zhang2015fixed}
Zhang, S., Jiang, H., Xu, M., Hou, J., and Dai, L.
\newblock The fixed-size ordinally-forgetting encoding method for neural
  network language models.
\newblock In \emph{Proceedings of Annual Meeting of the Association for
  Computational Linguistics (ACL)}, pp.\  495--500, 2015{\natexlab{d}}.

\end{thebibliography}
\bibliographystyle{icml2015}

 \end{document}